\documentclass{article}

\usepackage{microtype}
\usepackage{graphicx}
\usepackage{subfigure}
\usepackage{booktabs} % for professional tables

\usepackage{hyperref}

\usepackage[accepted]{icml2020}

\usepackage{graphicx}  %Required
\graphicspath{{images/}{plots/}}
\usepackage{pgfplots}
\pgfplotsset{compat=1.12}
\usepackage{amsmath}
\usepackage{multicol}
\usepackage{multirow}
\usepackage{booktabs}
\usepackage{color}
\usepackage{colortbl,array,xcolor}
\usepackage{silence}
\usepackage{amssymb} % for \smallsetminus
\usepackage{arydshln} % for \hdashline
\usepackage{CJKutf8}

\usepackage{soul}

\DeclareRobustCommand{\tyellow}[1]{\underline{{\textcolor{cyellow}{\bf #1}}}}

\DeclareRobustCommand{\hly}[1]{{\sethlcolor{cyellow}\hl{#1}}}

\DeclareMathOperator*{\argmax}{argmax} % no space, limits underneath in displays
\DeclareMathOperator*{\argandmax}{(arg)max} % no space, limits underneath in displays
\DeclareMathOperator*{\topk}{topk} % no space, limits underneath in displays

\usepackage{mathtools}

\DeclarePairedDelimiter\floor{\lfloor}{\rfloor}

\definecolor{bblue}{HTML}{4F81BD}
\definecolor{oorange}{HTML}{F4C842}
\definecolor{rred}{HTML}{C0504D}
\definecolor{ggreen}{HTML}{9BBB59}
\definecolor{ppurple}{HTML}{9F4C7C}
\definecolor{darkgreen}{HTML}{228B22}
\definecolor{cred}{HTML}{D81B60}
\definecolor{cblue}{HTML}{1E88E5}
\definecolor{cyellow}{HTML}{FFC107}
\definecolor{nred}{HTML}{e41a1c}
\definecolor{nblue}{HTML}{377eb8}
\definecolor{ngreen}{HTML}{4daf4a}
\definecolor{lblue}{HTML}{6C8EBF}

\newcommand{\RS}{{\tt RS}}
\newcommand{\Transformer}{{\tt Transformer}}
\newcommand{\Attention}{{\tt Attention}}
\newcommand{\DisCo}{{\tt DisCo}}
\newcommand{\Proj}{{\tt Proj}}
\newcommand{\obs}{{\tt obs}}
\newcommand{\mask}{{\tt mask}}
\newcommand{\Concat}{{\tt Concat}}

\icmltitlerunning{Non-autoregressive Machine Translation with Disentangled Context Transformer}
\begin{document}

\twocolumn[
\icmltitle{Non-autoregressive Machine Translation with Disentangled Context Transformer}
%\icmltitle{Parallel Machine Translation with Disentangled Context Transformer}

% It is OKAY to include author information, even for blind
% submissions: the style file will automatically remove it for you
% unless you've provided the [accepted] option to the icml2020
% package.

% List of affiliations: The first argument should be a (short)
% identifier you will use later to specify author affiliations
% Academic affiliations should list Department, University, City, Region, Country
% Industry affiliations should list Company, City, Region, Country

% You can specify symbols, otherwise they are numbered in order.
% Ideally, you should not use this facility. Affiliations will be numbered
% in order of appearance and this is the preferred way.
%\icmlsetsymbol{equal}{*}

\begin{icmlauthorlist}
\icmlauthor{Jungo Kasai}{uw}
\icmlauthor{James Cross}{fb}
\icmlauthor{Marjan Ghazvininejad}{fb}
\icmlauthor{Jiatao Gu}{fb}
\end{icmlauthorlist}

\icmlaffiliation{uw}{Paul G.\ Allen School of Computer Science \& Engineering, University of Washington. Work done at Facebook AI.}
\icmlaffiliation{fb}{Facebook AI}
\icmlcorrespondingauthor{Jungo Kasai}{jkasai@cs.washington.edu}

% You may provide any keywords that you
% find helpful for describing your paper; these are used to populate
% the "keywords" metadata in the PDF but will not be shown in the document
\icmlkeywords{Machine Learning, ICML}

\vskip 0.3in
]

% this must go after the closing bracket ] following \twocolumn[ ...

% This command actually creates the footnote in the first column
% listing the affiliations and the copyright notice.
% The command takes one argument, which is text to display at the start of the footnote.
% The \icmlEqualContribution command is standard text for equal contribution.
% Remove it (just {}) if you do not need this facility.

%\printAffiliationsAndNotice{}  % leave blank if no need to mention equal contribution
\printAffiliationsAndNotice{} % otherwise use the standard text.

\begin{abstract}
State-of-the-art neural machine translation models generate a translation from left to right and every step is conditioned on the previously generated tokens. The sequential nature of this generation process causes fundamental latency in inference since we cannot generate multiple tokens in each sentence in parallel.
We propose an attention-masking based model, called \textit{\textbf{Dis}entangled \textbf{Co}ntext} (DisCo) transformer, that simultaneously generates all tokens given different contexts.
The DisCo transformer is trained to predict every output token given an arbitrary subset of the other reference tokens.
We also develop the \textit{parallel easy-first} inference algorithm, which iteratively refines every token in parallel and reduces the number of required iterations.
Our extensive experiments on 7 translation directions with varying data sizes demonstrate that our model achieves competitive, if not better, performance compared to the state of the art in non-autoregressive machine translation while significantly reducing decoding time on average. Our code is available at \url{https://github.com/facebookresearch/DisCo}.
\end{abstract}

\section{Introduction}
\label{introduction}
State-of-the-art neural machine translation systems use \textit{autoregressive} decoding where words are predicted one-by-one conditioned on all previous words \cite{Bahdanau2014NeuralMT,Vaswani2017AttentionIA}.
\textit{Non-autoregressive} machine translation (NAT, \citealp{Gu2017NonAutoregressiveNM}), on the other hand, generates all words in one shot and speeds up decoding at the expense of performance drop.
Parallel decoding results in conditional independence and prevents the model from properly capturing the highly multimodal distribution of target translations \cite{Gu2017NonAutoregressiveNM}.
One way to remedy this fundamental problem is to refine model output iteratively \cite{Lee2018DeterministicNN, Ghazvininejad2019MaskPredictPD}. This work pursues this iterative approach to non-autoregressive translation.\footnote{Refinement requires several sequential steps, but we abuse the term \textit{non-autoregressive} generation to mean a broad family of methods that generate the target in parallel for simplicity.}

In this work, we propose a transformer-based architecture with attention masking, which we call \textit{\textbf{Dis}entangled \textbf{Co}ntext} (DisCo) transformer, and use it for non-autoregressive decoding.
Specifically, our DisCo transformer predicts every word in a sentence conditioned on an arbitrary subset of the rest of the words.
Unlike the masked language models \cite{devlins2019bert,Ghazvininejad2019MaskPredictPD} where the model only predicts the masked words, the DisCo transformer can predict all words simultaneously, leading to faster inference as well as a substantial performance gain when training data are relatively large.

We also introduce a new inference algorithm for iterative parallel decoding, \textit{parallel easy-first}, where each word is predicted by attending to the words that the model is more confident about.
This decoding algorithm allows for predicting all tokens with different contexts in each iteration and terminates when the output prediction converges, contrasting with the constant number of iterations \cite{Ghazvininejad2019MaskPredictPD}.
Indeed, we will show in a later section that this method substantially reduces the number of required iterations without loss in performance.
%Our decoder also substantially reduces the amount of decoding computation necessary by making crucial use of contextless information.
%Concretely, all key and value vectors in the transformer are projections from contextless word embeddings and only query vectors are contextual.
%This allows for reusing key and value vectors regardless of the context.

Our extensive empirical evaluations on 7 translation directions from standard WMT benchmarks show that our approach achieves competitive performance to state-of-the-art non-autoregressive and autoregressive machine translation while significantly reducing decoding time on average.

\section{DisCo Transformer}
\begin{figure}[h]
    \centering
    \includegraphics[width=0.46\textwidth]{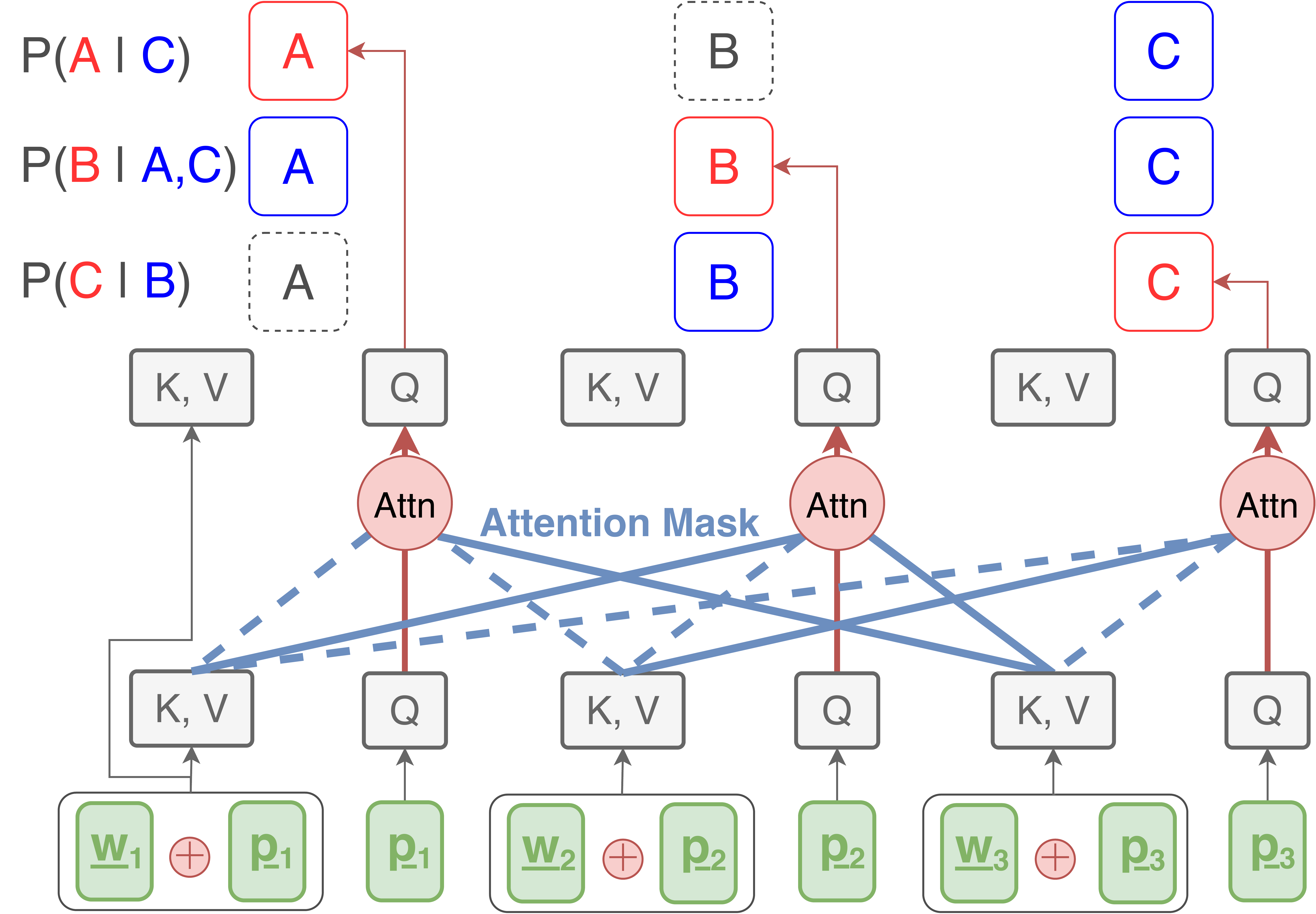}
    \caption{DisCo Transformer. $W$ and $p$ denote word and positional embeddings respectively. We simulate three \textit{disentangled} contexts to predict \textcolor{red}{A}, \textcolor{red}{B}, and \textcolor{red}{C} ($Y_n$) given \{\textcolor{blue}{C}\}, \{\textcolor{blue}{A}, \textcolor{blue}{C}\}, and \{\textcolor{blue}{B}\} ($Y_{\obs}^n$) respectively. \textcolor{lblue}{\bf Dashed lines} indicate masked-out attention connections to $Y_{\mask}^n$ and $Y_n$ itself. $K$ and $V$ are direct projections of $w_n+p_n$ (thus contextless) for all layers to avoid leakage.} 
    \label{fig:disco}
\end{figure}
We introduce our DisCo transformer for non-autoregressive translation (Fig.\ \ref{fig:disco}).
We propose a DisCo objective as an efficient alternative to masked language modeling and design an architecture that computes the objective in a single pass.
\subsection{DisCo Objective}
\label{subsec:generalization}
Similar to masked language models \cite{devlins2019bert}, a conditional masked language model (CMLM, \citealp{Ghazvininejad2019MaskPredictPD})
predicts randomly masked target tokens $Y_{\mask}$ given a source text $X$ and the rest of the target tokens $Y_{\obs}$. Namely, for every sentence pair in bitext $X$ and $Y$, 
\begin{align*}
P(Y_{\mask}| X, Y_{\obs}) = \Transformer(X, Y_{\obs})\\
Y_{\mask} \sim \RS(Y) \quad Y_{\obs} = Y \setminus Y_{\tt mask}
\end{align*}
where $\RS$ denotes random sampling of masked tokens.\footnote{BERT \cite{devlins2019bert} masks a token with probability 0.15 while CMLMs \cite{Ghazvininejad2019MaskPredictPD} sample the number of masked tokens uniformly from $[1, N]$.}
CMLMs have proven successful in parallel decoding for machine translation \cite{Ghazvininejad2019MaskPredictPD}, video captioning \cite{Video2019}, and speech recognition \cite{Speech2019}.
However, the fundamental inefficiency with this masked language modeling objective is that the model can only be trained to predict a subset of the reference tokens ($Y_{\mask}$) for each network pass unlike a normal autoregressive model where we predict all $Y$ from left to right. 
%Indeed, we will show in a later section that this problem becomes particularly severe when the training bitext is relative large.
To address this limitation, we propose a \textit{\textbf{Dis}entangled \textbf{Co}ntext} (DisCo) objective.\footnote{We distinguish this from disentangled representation.}
The objective involves prediction of every token given an arbitrary (thus \textit{disentangled}) subset of the other tokens. For every $1\leq n \leq N$ where $|Y| = N$, we predict:
\begin{align*}
P(Y_n | X, Y_{\obs}^n) = \Transformer(X, Y_{\obs}^n)\\
Y_{\obs}^n \sim \RS(Y\setminus Y_n) 
\end{align*}

\subsection{DisCo Transformer Architecture}
Simply computing conditional probabilities $P(Y_n|X, Y_{\obs}^n)$ with a vanilla transformer decoder will necessitate $N$ separate transformer passes for each $Y_{\obs}^n$.
We introduce the DisCo transformer to compute these $N$ \textit{contexts} in one shot:
\begin{align*}
P(Y_1| X, Y_{\obs}^1),\cdots, P(Y_N| X, Y_{\obs}^N) = \DisCo(X, Y)
\end{align*}
In particular, our DisCo transformer makes crucial use of attention masking to achieve this computational efficiency.
Denote input word and positional embeddings at position $n$ by $w_n$ and $p_n$.
For each position $n$ in $Y$, the vanilla transformer computes self-attention:\footnote{For simplicity, here we omit fully-connected layers, layer-norm, residual connections, and cross attention to the encoder.}
\begin{align*}
    &k_n, v_n, q_n = \Proj(w_n + p_n)\\
    &h_n = \Attention(K, V, q_{n})\\
    &K, V = \Concat\left(\{k_m\}_{m=1}^N\right), \Concat\left(\{v_m\}_{m=1}^N\right)
    %\text{where } K =\Concat \left( \left\{ k_m | 1 \leq m \leq N  \right\} \right)\\
    %V = \Concat \left( \left\{ v_m | 1 \leq m \leq N  \right\} \right)
\end{align*}
%where $K$ and $V$ denote concatenated matricies of $k_n$ and $v_n$ for $1 \leq n \leq N$.
%where $K, V = \Concat\left(\{k_m\}_{m=1}^N\right), \Concat\left(\{v_m\}_{m=1}^N\right)$.
We modify this attention computation in two aspects. 
First, we separate query input from key and value input to avoid feeding the token we predict.
Then we only attend to keys and values that correspond to observed tokens ($K_{\obs}^n$, $V_{\obs}^n$) and \textit{mask out} the connection to the other tokens ($Y_{\mask}^n$ and $Y_n$ itself, \textcolor{lblue}{\bf dashed lines} in Fig.\ \ref{fig:disco}). 
\begin{align*}
    &k_n, v_n = \Proj(w_n + p_n) \quad q_n = \Proj(p_n)\\
    &h_n = \Attention(K_{\obs}^n, V_{\obs}^n, q_{n})\\
    &K_{\obs}^n =\Concat \left( \left\{ k_m | Y_m \in Y_{\obs}^n  \right\} \right)\\
    &V_{\obs}^n =\Concat \left( \left\{ v_m | Y_m \in Y_{\obs}^n  \right\} \right)
\end{align*}

\subsection{Stacked DisCo Transformer}
Unfortunately stacking DisCo transformer layers is not straightforward.
Suppose that we compute the $n$th position in the $j$th layer from the prevous layer's output as follows:
\begin{align*}
    &k_n^j, v_n^j = \Proj(w_n + h_n^{j-1}) \quad q_n^{j} = \Proj(h_n^{j-1})\\
    &h_n^j = \Attention(K_{\obs}^{n, j}, V_{\obs}^{n, j}, q_{n}^j)
\end{align*}
In this case, however, any cyclic relation between positions will cause information leakage.
Concretely, assume that $Y = [A, B]$ and $N=2$.
Suppose also that $Y_{obs}^1 = B$ and $Y_{obs}^2=A$, and thus there is a cycle that position 1 can see $B$ and position 2 can see $A$.
Then the output state at position 1 in the first layer $h_1^1$  becomes a function of $B$:
\begin{align*}
h_1^1 (B) = \Attention(k_2^1(B), v_2^1(B), q_1^1)
\end{align*}
Since position 2 can see position 1, the output state at position 2 in the second layer $h_2^2$ is computed by
\begin{align*}
h_2^2 = \Attention\left(k_1^2(h_1^1(B)), v_1^2(h_1^1(B)), q_2^2\right)
\end{align*}
But $h_2^2$ will be used to predict the token at position 2 i.e., $B$, and this will clearly make the prediction problem degenerate. 
To avoid this cyclic leakage, we make keys and values independent of the previous layer's output $h_n^{j-1}$:
\begin{align*}
    &k_n^j, v_n^j = \Proj(w_n + p_n) \quad q_n^j = \Proj(h_n^{j-1})\\
    &h_n^j = \Attention(K_{\obs}^{n,j}, V_{\obs}^{n,j}, q_{n}^j)
\end{align*}
In other words, we \textit{decontextualize} keys and values in stacked DisCo layers.

\subsection{Training Loss}
We use a standard transformer as an encoder and stacked DisCo layers as a decoder.
For each $Y_n$ in $Y$ where $|Y|=N$, we uniformly sample the number of visible tokens from $[0, N-1]$, and then we randomly choose that number of tokens from $Y\setminus Y_n$ as $Y_{\obs}^n$, similarly to CMLMs \cite{Ghazvininejad2019MaskPredictPD}.
We optimize the negative log likelihood loss from $P(Y_n | X, Y_{\obs}^n)$ $(1 \leq n \leq N)$.
Again following CMLMs, we append a special token to the encoder and project the vector to predict the target length for parallel decoding.
We add the negative log likelihood loss from this length prediction to the loss from word predictions.

\subsection{DisCo Objective as Generalization}
We designed the DisCo transformer to compute conditional probabilities at every position efficiently, but here we note that the DisCo transformer can be readily used with other training schemes in the literature.
We can train an autoregressive DisCo transformer by always setting $Y_{\obs}^n = Y_{<n}$.
XLNet \cite{Yang2019XLNetGA} is also a related variant of a transformer that was introduced to produce general-purpose contextual word representations.
The DisCo transformer differs from XLNet in two critical ways. First, XLNet consists of separate context stream and query stream attention. This means that we need to double the amount of expensive attention and fully connected layer computation in the transformer.
Another difference is that XLNet is only trained to predict tokens in permuted order.
The DisCo transformer can be trained for the permutation objective by setting $Y_{\obs}^n = \{ Y_i  | z(i) <z(n)\}$ where $z(i)$ indicates the rank of the $i$th element in the new order.
\citet{Baevski2019ClozedrivenPO} train their two tower model with the \textit{cloze objective} again for general-purpose pretraining.
We can train our DisCo transformer with this objective by setting $Y_{\obs}^n = \{ Y_{\neq n}\}$.
The proposed DisCo objective provides generalization that encompasses all of these special cases.
%Our DisCo transformer provides generalization that encompasses all of these special cases.
%Later we will demonstrate that our most generalized multimask objective performs best in our experiments.
\section{Inference Algorithms}
In this section, we discuss inference algorithms for our DisCo transformer.
We first review \textit{mask-predict} from prior work as a baseline and introduce a new parallelizable inference algorithm, \textit{parallel easy-first} (Alg.\ \ref{alg:parallel_eas_first}).

\subsection{Mask-Predict}
\textit{Mask-predict} is an iterative inference algorithm introduced in \citet{Ghazvininejad2019MaskPredictPD} to decode a conditional masked language model (CMLM).
The target length $N$ is first predicted, and then the algorithm iterates over two steps: \textit{mask} where $i_t$ tokens with lowest probability are masked and \textit{predict} where those masked tokens are updated given the other $N-i_t$ tokens.
The number of masked tokens $i_t$ decays from $N$ with a constant rate over a fixed number of iterations $T$. Specifically, at iteration $t$,
\begin{align*}
    &\quad i_t = \floor*{N \cdot \frac{T-t+1}{T}}\\
    &Y_{\obs}^t = \{Y_{j}^{t-1} | j \in \topk_{n} (p_n^{t-1}, k=N-i_t) \}\\
    &Y_n^t,p_n^t =
    \begin{cases}
    \argandmax_{w} P(Y_n=w | X, Y_{\obs}^t) \ \text{if } Y_n^t \not\in Y_{\obs}^t\\
        Y_n^{t-1}, p_n^{t-1} \quad \quad \text{otherwise}
    \end{cases}
\end{align*}
This method is directly applicable to our DisCo transformer by fixing $Y_{obs}^{n,t}$ regardless of the position $n$.

\subsection{Parallel Easy-First}
\label{sec:easy-first}
An advantage of the DisCo transformer over a CMLM is that we can predict tokens in all positions conditioned on different context simultaneously.
The mask-predict inference can only update masked tokens given the fixed observed tokens $Y_{\obs}^t$, meaning that we are wasting the opportunity to improve upon $Y_{\obs}^t$ and to take advantage of broader context present in $Y_{\mask}^t$.
We develop an algorithm, \textit{parallel easy-first} (Alg.\ \ref{alg:parallel_eas_first}), which makes predictions in all positions, thereby benefiting from this property.
Concretely, in the first iteration, we predict all tokens in parallel given source text:
\begin{align*}
 Y_n^1, p_n = \argandmax_w P(Y_n=w | X) 
\end{align*}
Then, we get the \textit{easy-first} order $z$ where $z(i)$ denotes the rank of $p_i$ in descending order.
At iteration $t>1$, we update predictions for all positions by
\begin{align*}
   &Y_{\obs}^{n, t} = \left\{ Y_i^{t-1} | z(i) < z(n) \right\}\\
   &Y_n^t, p_n^t = \argandmax_{w} P\left(Y_n=w | X, Y_{\obs}^{n,t}\right)
\end{align*}
Namely, we update each position given previous predictions on the \textit{easier} positions.
In a later section, we will explore several variants of choosing $Y_{\obs}^{n,t}$ and show that this easy-first strategy performs best despite its simplicity.% and succeeds in reducing iterations while keeping performance.

\subsection{Length Beam}
Following \citet{Ghazvininejad2019MaskPredictPD}, we apply length beam. 
In particular, we predict top K lengths from the distribution in length prediction and run parallel easy-first simultaneously.
In order to speed up decoding, we terminate if the one with the highest average log score $\sum_{n=1}^N \log(p_n^t)/N$ converges.
It should be noted that for parallel easy-first, $Y^t = Y^{t-1}$ means convergence because $Y^{n,t}_{obs}=Y^{n,t+1}_{obs}$ for all positions $n$ while mask-predict may keep updating tokens even after because $Y_{\obs}^t$ changes over iterations.
See Alg.\ \ref{alg:parallel_eas_first} for full pseudo-code.
Notice that all for-loops are parallelizable except the one over iterations $t$.
In the subsequent experiments, we use length beam size of 5 \cite{Ghazvininejad2019MaskPredictPD} unless otherwise noted.
In Sec.\ \ref{sec:analysis_inference}, we will illustrate that length beam facilitates decoding both the CMLM and DisCo transformer.

\begin{algorithm}[tb]
%\footnotesize
\caption{Parallel Easy-First with Length Beam}
\label{alg:parallel_eas_first}
\begin{algorithmic}
%\REQUIRE ~~\\
\STATE Source sentence: $X$\\
\STATE Predicted lengths: $N_1, \cdots , N_K$\\
\STATE Max number of iterations: $T$
\FOR{ { $k \in \left\{1,2,...,K\right\}$}}
\FOR{ { $n \in \left\{1,2,...,N_k\right\}$}}
    \STATE {$Y^{1,k}_n, p_n^k = \argandmax_{w} P (y_n=w| X)$}
\ENDFOR
\STATE{Get the easy-first order $z_k$ by sorting $p^k$ and let $z_k(i)$ be the rank of the $i$th position.}
\ENDFOR
\FOR{ { $t \in \left\{2,...,T\right\}$}}
\FOR{ { $k \in \left\{1,...,K\right\}$}}
\FOR{{$n \in \left\{1,2,...,N_{k}\right\}$}}
\STATE{
\vspace{-0.3cm}
\begin{align*}
   Y_{\obs}^{n, t} = \{ Y_i^{t-1, k} | z_k(i) < z_k(n) \}\\
   Y_n^{t,k}, p_n^{t,k} = \argandmax_{w} P\left(Y_n=w | X, Y_{\obs}^{n,t}\right)
\end{align*}
\vspace{-0.5cm}
}
\ENDFOR
\ENDFOR
\STATE{
\vspace{-0.8cm}
\begin{align*}
   k^* = \argmax_k \sum_{n=1}^{N_k} \log\left(p_n^{t,k} \right)/N_k
\end{align*}
}
%\IF{$Y^{t-1,k^*} = Y^{t,k^*}$} 
%\STATE {\textbf{return} $Y^{t,k^*}$}
\STATE {\textbf{if} $Y^{t-1,k^*} = Y^{t,k^*}$ \textbf{then} \textbf{return} $Y^{t,k^*}$}
%\ENDIF
\ENDFOR
\STATE {\textbf{return} $Y^{T, k^*}$}
\end{algorithmic}
\end{algorithm}

\section{Experiments}
We conduct extensive experiments on standard machine translation benchmarks.
We demonstrate that our DisCo transformer with the parallel easy-first inference achieves comparable performance to, if not better than, prior work on non-autoregressive machine translation with substantial reduction in the number of sequential steps of transformer computation.
We also find that our DisCo transformer achieves more pronounced improvement when bitext training data are large, getting close to the performance of autoregressive models.

\subsection{Experimental Setup}
\label{sec:setup}
\paragraph{Benchmark datasets}
We evaluate on 7 directions from four standard datasets with various training data sizes: WMT14 EN-DE (4.5M pairs), WMT16 EN-RO (610K pairs), WMT17 EN-ZH (20M pairs), and WMT14 EN-FR (36M pairs, en$\rightarrow$fr only).
These datasets are all encoded into subword units by BPE \cite{sennrich-etal-2016-neural}.\footnote{We run joint BPE on all language pairs except EN-ZH.}
We use the same preprocessed data and train/dev/test splits as prior work for fair comparisons (EN-DE: \citealp{Vaswani2017AttentionIA}; EN-RO: \citealp{Lee2018DeterministicNN}; EN-ZH: \citealp{Hassan2018AchievingHP, wu2018pay}; EN-FR: \citealp{Gehring2017ConvolutionalST, Ott2018ScalingNM}).
We evaluate performance with BLEU scores \cite{Papineni2001BleuAM} for all directions except that we use SacreBLEU \cite{post-2018-call}\footnote{SacreBLEU hash: BLEU+case.mixed+lang.en-zh+numrefs.1+smooth.exp+test.wmt17+tok.zh+version.1.3.7.} in en$\rightarrow$zh again for fair comparison with prior work \cite{Ghazvininejad2019MaskPredictPD}.
For all autoregressive models, we apply beam search with $b=5$ \cite{Vaswani2017AttentionIA, Ott2018ScalingNM} and tune length penalty of $\alpha \in [0.0, 0.2, \cdots, 2.0]$ in validation.
For parallel easy-first, we set the max number of iterations $T=10$ and use $T=4,10$ for constant-time mask-predict.

\begin{table*}[t]
\centering
%\small
\caption{The performance of non-autoregressive machine translation methods on the WMT14 EN-DE and WMT16 EN-RO test data. The Step columns indicate the average number of sequential transformer passes.
Shaded results use a small transformer ($d_{model}=d_{hidden}=512$).  
Our EN-DE results show the scores after conventional compound splitting \cite{luong-etal-2015-effective, Vaswani2017AttentionIA}.}
\vskip 0.1in
%\addtolength{\tabcolsep}{-1.5pt}  
\begin{tabular}{l|cc|cc|cc|cc}
\toprule
%& \textbf{I} & \multicolumn{2}{c}{\textbf{WMT'14}}  \\
\textbf{Model}&
\multicolumn{2}{c}{\textbf{en$\rightarrow$de}} &  
\multicolumn{2}{c|}{\textbf{de$\rightarrow$en}}  & 
\multicolumn{2}{c}{\textbf{en$\rightarrow$ro}} &  
\multicolumn{2}{c}{\textbf{ro$\rightarrow$en}} \\
$n$: \# rescored candidates &  Step\ & BLEU& Step & BLEU & Step & BLEU & Step & BLEU \\
\hline
\rowcolor[gray]{.90} \citet{Gu2017NonAutoregressiveNM} ($n=100$)& 1 & 19.17 & 1 & 23.20  & 1 &29.79 &1 & 31.44 \\
\citet{Wang2019NonAutoregressiveMT} ($n=9$) & 1 & 24.61 & 1 & 28.90  & --& --& --&-- \\
\citet{Li2019HintBasedTF} ($n=9$)& 1 & 25.20 & 1 &  28.80  & -- & -- & -- & --\\
\citet{Ma2019FlowSeqNC} ($n=30$) & 1 & 25.31 & 1 & 30.68 & 1 & 32.35 & 1 & 32.91\\
\citet{Sun2019Fast} ($n=19$) & 1 & 26.80  & 1 &  30.04 & -- & -- & -- & -- \\
\rowcolor[gray]{.90} \citet{ReorderNAT} & 1 & 26.51& 1 &  31.13 & 1 & 31.70 & 1 & 31.99\\
\citet{Shu2019LatentVariableNN} ($n=50$) & 1 & 25.1 & -- & -- & -- & -- & -- & --\\ \hline
\textbf{Iterative NAT Models} & & & & & & & &\\
%\citet{Libovick2018EndtoEndNN} & 1+ & 17.68 & 1 & 19.80\\
\rowcolor[gray]{.90} \citet{Lee2018DeterministicNN}  & 10 & 21.61 & 10 & 25.48 & 10 & 29.32 & 10 & 30.19 \\
\citet{Ghazvininejad2019MaskPredictPD} (CMLM) & 4& 25.94 & 4 & 29.90 & 4 &  32.53 & 4 & 33.23 \\
 & 10& 27.03 & 10 & 30.53 &10 & 33.08&10 & 33.31\\
%\citet{Stern2019InsertionTF} & $\log N$& 27.41& --& --\\
\citet{Gu2019LevenshteinT} (LevT) & 7+ & 27.27 &--  & -- & -- & -- &  7+ & 33.26 \\
\hdashline
%CMLM + Mask-Predict & 5 &26.42 (27.03)&4& 30.33 (30.75) & 4 & 33.02 & 3 & 33.24\\
\textbf{Our Implementations} & & & & & & & &\\
CMLM + Mask-Predict & 4 & 26.73&4& 30.75 & 4 & 33.02 & 4 &  33.27\\
CMLM + Mask-Predict & 10 & \textbf{27.39} & 10 & 31.24 & 10 &  \textbf{33.33} & 10 &  \textbf{33.67} \\
%Levenshtein Tranformer & & & & &&&\\
%Multimask + Mask-Predict & 5& 25.75 (26.45) & 5& 30.11 (30.53) & 4 &32.22 & 3 & 32.64\\
DisCo + Mask-Predict & 4& 25.83 & 4& 30.15 & 4 &32.22 & 4& 32.92\\
%Multimask + Mask-Predict & 5& 25.75 (26.45) & 5& 30.11 (30.53) & 4 &32.63 & 3 & 32.92 \\
%Multimask + Mask-Predict & 10& 26.45 (27.06) &10&30.48 (30.89) & 10 & 32.44& 10 &32.86\\
DisCo + Mask-Predict & 10& 27.06 &10&30.89 & 10 & 32.92 & 10 & 33.12\\
%is Multimask + Easy-First & 4.82 & 26.75 (27.34)& 4.23 & 30.89 (\textbf{31.31}) &3.29 &32.73 &3.09 & 33.14 \\ 
DisCo + Easy-First & 4.82 & 27.34& 4.23 & \textbf{31.31} &3.29 &33.22 &3.10 & 33.25 \\ 
\hline 
\textbf{AT Models} & & & & & & & &\\
\citet{Vaswani2017AttentionIA} (base)& N&27.3 &--& -- & -- & -- & -- & --\\
\citet{Vaswani2017AttentionIA} (large) & N &28.4 &--& -- & -- & -- & -- & --\\
\hdashline
\textbf{Our Implementations} & & & & & & & &\\
AT Transformer Base (EN-RO teacher) & N & 27.38 & N  & 31.78& N& 34.16 & N & 34.46\\
AT Transformer Base + Distillation\ & N & 28.24 & N  & 31.54 & --& -- & -- & --\\
AT Transformer Large (EN-DE teacher) & N & 28.60 & N & 31.71 & -- & -- & -- & --\\
\bottomrule
\end{tabular}
\label{en-de-ro_result}
\vskip -0.1in
\end{table*}

\subsection{Baselines and Comparison}
There has been a flurry of recent work on non-autoregressive machine translation (NAT) that finds a balance between parallelism and performance.
Performance can be measured using automatic evaluation such as BLEU scores \cite{Papineni2001BleuAM}.
Latency is, however, challenging to compare across different methods.
For models that have an autoregressive component (e.g.,\ \citealp{Kaiser2018FastDI,ReorderNAT}), we can speed up sequential computation by caching states.
%\citet{Ma2019FlowSeqNC} demonstrated that their flow-based model benefits more from large batches as compared to a standard autoregressive model.
Further, many of prior NAT approaches generate varying numbers of translation candidates and rescore them using an autoregressive model.
The rescoring process typically costs overhead of one parallel pass of a transformer encoder followed by a decoder.
Given this complexity in latency comparison, we highlight two state-of-the-art iteration-based NAT models whose latency is comparable to our DisCo transformer due to the similar model structure.
See Sec.\ \ref{sec:related_work} for descriptions of more work on NAT.
\paragraph{CMLM}
As discussed earlier, we can generate a translation with mask-predict from a CMLM \cite{Ghazvininejad2019MaskPredictPD}.
We can directly compare our DisCo transformer with this method by the number of iterations required.\footnote{Caching \textit{contextless} key and value computation in the DisCo transformer gives us a slight speedup, but it is relatively minor as compared to expensive attention and fully connected computation.}
%We also use length beam size of 5 to ensure comparability in all experiments.
We provide results obtained by running their code.\footnote{\url{https://github.com/facebookresearch/Mask-Predict}}

\paragraph{Levenshtein Transformer}
Levenshtein transformer (LevT) is a transformer-based iterative model for parallel sequence generation \cite{Gu2019LevenshteinT}.
Its iteration consists of three sequential steps: \textit{deletion}, \textit{placeholder prediction}, and \textit{token prediction}.
%While Levenshtein transformer has three times more decoder parameters than the multimask transformer and CMLMs
Unlike the CMLM with the constant-time mask-predict inference, decoding in LevT terminates adaptively under certain condition.
Its latency is roughly comparable by the average number of sequential transformer runs.
Each iteration consists of three transformer runs except that the first iteration skips the \textit{deletion} step.
%For en$\leftrightarrow$zh, and en$\rightarrow$fr, we use their code\footnote{\url{https://github.com/pytorch/fairseq/tree/master/examples/nonautoregressive_translation}} and report the average number of these sequential steps for comparison.
See \citet{Gu2019LevenshteinT} for detail.

\paragraph{Hyperparameters}
We generally follow the hyperparameters for a transformer base \cite{Vaswani2017AttentionIA, Ghazvininejad2019MaskPredictPD}: 6 layers for both the encoder and decoder, 8 attention heads, 512 model dimensions, and 2048 hidden dimensions.
We sample weights from $\mathcal{N}(0, 0.02)$, initialize biases to zero, and set layer normalization parameters to $\beta=0,\gamma=1$ \cite{devlins2019bert}.
For regularization, we tune the dropout rate from  $[0.1, 0.2, 0.3]$ based on dev performance in each direction, and apply weight decay with $0.01$ and label smoothing with $\varepsilon=0.1$.
We train batches of approximately 128K tokens using Adam \cite{Kingma2014AdamAM} with $\beta=(0.9, 0.999)$ and $\varepsilon=10^{-6}$.
The learning rate warms up to $5\cdot10^{-4}$ in the first 10K steps, and then decays with the inverse square-root schedule.
We train all models for 300K steps apart from en$\rightarrow$fr where we make 500K steps to account for the data size.
We measure the dev BLEU score at the end of each epoch to avoid stochasticity, and average the 5 best checkpoints to obtain the final model.
We use 16 Telsa V100 GPUs and accelerate training by mixed precision floating point \cite{micikevicius2018mixed}, and implement all models with \texttt{fairseq} \cite{ott-etal-2019-fairseq}.

%\vspace{-0.5cm}
\paragraph{Distillation}
Similar to previous work on non-autoregressive translation (e.g.,\ \citealp{Gu2017NonAutoregressiveNM, Lee2018DeterministicNN}), we apply sequence-level knowledge distillation \cite{Kim2016SequenceLevelKD} by training every model in all directions on translations produced by a standard left-to-right transformer model (transformer large for EN-DE, EN-ZH, and EN-FR and base for EN-RO).
We also present results obtained from training a standard autoregressive base transformer on the same distillation data for comparison.
%to ensure comparability. 
We assess the impact of distillation in Sec.\ \ref{sec:analysis_training} and demonstrate that distillation is still a key component in our non-autoregressive models.

\subsection{Results and Discussion}
Seen in Table \ref{en-de-ro_result} are the results in the four directions from the WMT14 EN-DE and WMT16 EN-RO datasets.
First, our re-implementations of CMLM + Mask-Predict outperform \citet{Ghazvininejad2019MaskPredictPD} (e.g.,\ 31.24 vs.\ 30.53 in de$\rightarrow$en with 10 steps). This is probably due to our tuning on the dropout rate and weight averaging of the 5 best epochs based on the validation BLEU performance (Sec.\ \ref{sec:setup}).

Our DisCo transformer with the parallel easy-first inference achieves at least comparable performance to the CMLM with 10 steps despite the significantly fewer steps on average (e.g.,\ 4.82 steps in en$\rightarrow$de).
The one exception is ro$\rightarrow$en (33.25 vs. 33.67), but DisCo + Easy-First requires only 3.10 steps, and CMLM + Mask-Predict with 4 steps achieves similar performance of 33.27.
The limited advantage of our DisCo transformer on the EN-RO dataset suggests that we benefit less from the training efficiency of the DisCo transformer on the small dataset (610K sentence pairs).
DisCo + Mask-Predict generally underperforms DisCo + Easy-First, implying that the mask-predict inference, which fixes $Y_{\obs}^n$ across all positions $n$, fails to utilize the flexibility of the DisCo transformer.
DisCo + Easy-First also accomplishes significant reduction in the average number of steps as compared to the adaptive decoding in LevT \cite{Gu2019LevenshteinT} while performing competitively.
As discussed earlier, each iteration in inference on LevT involves three sequential transformer runs, which undermine the latency improvement.

Overall, our implementations compare well with other NAT models from prior work.
We achieve competitive performance to the standard autoregressive models with the same transformer base configuration on the EN-DE dataset except that the autoregressive model with distillation performs comparably to the transformer large teacher in en$\rightarrow$de (28.24 vs. 28.60).
Nonetheless, we still see a large gap between the autoregressive teachers and our NAT results in both directions from EN-RO, illustrating a limitation of our remedy for the trade-off between decoding parallelism and performance.

Tables \ref{en-zh_result} and \ref{en-fr_result} show results from the EN-ZH and EN-FR datasets where the bitext data are larger (20M and 36M sentence pairs).
In both cases we see similar (yet more pronounced) patterns to the EN-DE and EN-RO experiments. Particularly noteworthy is that DisCo with the parallel easy-first inference and dropout tuning yields 34.63 points, a gain of 1.4 BLEU improvement over \citet{Ghazvininejad2019MaskPredictPD} in en$\rightarrow$zh despite the average of 5.44 steps.

\begin{table}[h]
\centering
%\small
\caption{WMT17 EN-ZH test results.}
\vskip 0.1in
\addtolength{\tabcolsep}{-1.5pt}  
\begin{tabular}{l|cc|cc}
\toprule
%& \textbf{I} & \multicolumn{2}{c}{\textbf{WMT'14}}  \\
\textbf{Model}& \multicolumn{2}{c|}{\textbf{en$\rightarrow$zh}} &  \multicolumn{2}{c}{\textbf{zh$\rightarrow$en}}  \\
& Step\ & BLEU& Step\ & BLEU\\
\hline
\citeauthor{Ghazvininejad2019MaskPredictPD} & 4& 32.63 & 4 & 21.90  \\
(\citeyear{Ghazvininejad2019MaskPredictPD}) & 10& 33.19 & 10 & 23.21  \\
 \hdashline
%\hline
%multimask + Mask-Predict & 10&26.92&10& \\
%Levenshtein Transformer \\
\textbf{Our Implementations} & & & & \\
CMLM + Mask-Predict & 4 & 33.58 & 4 &  22.57\\
CMLM + Mask-Predict & 10 &  34.24 & 10 & 23.76\\
DisCo + Mask-Predict & 4 & 33.61 &4 & 22.42\\ 
DisCo + Mask-Predict & 10 & 34.51 &10 & 23.68\\ 
DisCo + Easy-First & 5.44 &\textbf{34.63} &5.90 & \textbf{23.83}\\ 
\hline 
AT Transformer Base & N & 34.74 & N & 23.77\\
+ Distillation& N & 35.09 & N & 24.53\\
AT Trans.\ Large (teacher)& N & 35.01 & N & 24.65\\
%& 4 & \textbf{25.94} & \textbf{29.90}  \\
%\midrule
%& 10 & \textbf{27.03} & \textbf{30.53}  \\
%\citet{Gu2019LevenshteinT} 23.76& &  &   \\
%& 4 & \textbf{25.94} & \textbf{29.90}  \\
%\midrule
%Base Transformer \cite{vaswani2017} &  $N$~~~~~~ & 27.30 & ---~---  \\
%Base Transformer (Our Implementation) & $N$~~~~~~ & 27.74 & 31.09  \\
%Base Transformer (+Distillation) &  $N$~~~~~~ & 27.86 & 31.07  \\
%Large Transformer \cite{vaswani2017} & $N$~~~~~~ & 28.40 & ---~---  \\
%Large Transformer (Our Implementation) &  $N$~~~~~~ & 28.60 & 31.71 \\
\bottomrule
\end{tabular}
\label{en-zh_result}
\end{table}

\begin{table}[h]
\centering
%\small
\caption{WMT14 EN-FR test results.}
\vskip 0.1in
\begin{tabular}{l|ccc}
\toprule
\textbf{Model} & \multicolumn{2}{c}{\textbf{en$\rightarrow$fr}} & Train  \\
&  Step & BLEU & Time \\
%Levenshtein Transformer & 7.74 & \textbf{40.94} & 106 h\\
\hline
CMLM + Mask-Predict & 4 & 40.21 & \multirow{2}{*}{53 h}\\
%CMLM + Mask-Predict & 5 & 40.38 & \multirow{2}{*}{53 h}\\
CMLM + Mask-Predict & 10&  40.55 & \\
\hdashline
DisCo + Mask-Predict & 4 &  39.59 & \multirow{3}{*}{37 h}\\
DisCo + Mask-Predict & 10 & 40.27& \\
DisCo + Easy-First & 4.29 &  40.66 & \\
%Multimask Large + Mask-Predict & 10 \\
%Multimask Large + Easy-First & & \\ 
\hline
\citet{Vaswani2017AttentionIA} (base) & N & 38.1 & --\\
\citet{Vaswani2017AttentionIA} (large) & N & 41.8 & --\\
\citet{Ott2018ScalingNM} (teacher) & N & 43.2 & --\\\hdashline
AT Transformer Base & N & 41.27 & 28 h\\
+ Distillation & N &42.03 & 28 h\\
\bottomrule
\end{tabular}
\label{en-fr_result}
\end{table}

\subsection{Decoding Speed}
\begin{figure}[t]
\centering
    \includegraphics[width=0.46\textwidth]{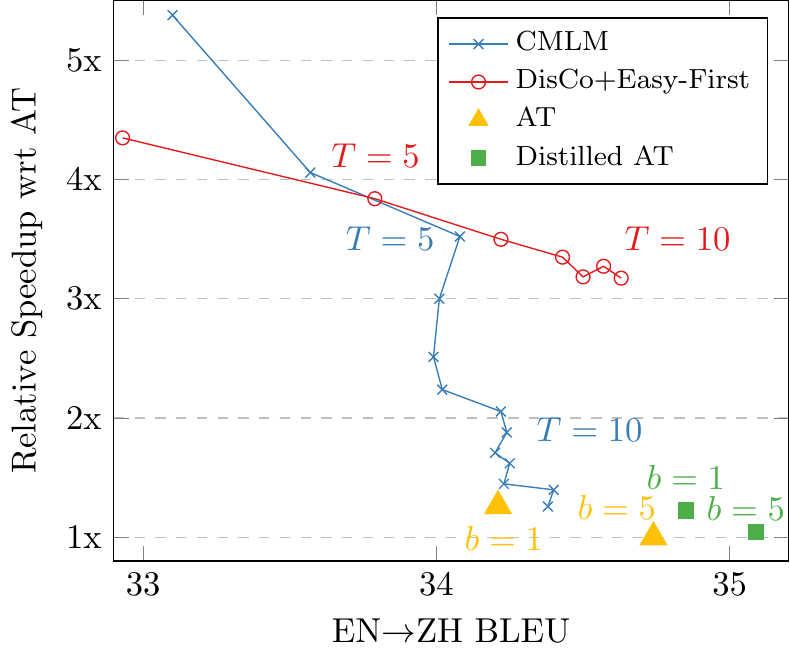}
\caption{Relative decoding speedup on the en$\rightarrow$zh test data with respect to the standard autoregressive model (indicated as \textcolor{cyellow}{$\blacktriangle$}). $T$ and $b$ denote the (max) number of iterations and beam size respectively. The length beam size is all set to 5.}
\label{fig:en-zh_speedup}
\vspace{-0.5cm}
\end{figure}

We saw the the DisCo transformer with the parallel easy-first inference achieves competitive performance to the CMLM while reducing the number of iterations.
Here we compare them in terms of the wall-time speedup with respect to the standard autoregressive model of the same base configuration (Fig.\ \ref{fig:en-zh_speedup}).
For each decoding run, we feed one sentence at a time and measure the wall time from when the model is loaded until the last sentence is translated, following the setting in \citet{Gu2019LevenshteinT}.
All models are implemented in \texttt{fairseq} \cite{ott-etal-2019-fairseq} and run on a single Nvidia V100 GPU.
%We run the CMLM with varying number of iterations $T$ to see the trade-off between performance and parallelism in constant-time decoding.
We can confirm that the average number of iterations directly translates to decoding time; the average number of iterations of the DisCo transformer with $T=10$ was 5.44 and the measured speedup lies between $T=5,6$ of the CMLM.
Note that \texttt{fairseq} implements effcient decoding of autoregressive models by caching hidden states.
The average length of generated sentences in the autoregressive model was 25.16 (4.6x steps compared to 5.44 steps), but we only gained a threefold speedup from DisCo.

\section{Analysis and Ablations}
%We presented several techniques to improve performance in parallel decoding.
In this section, we give an extensive analysis on our apporach along training and inference dimensions.

\subsection{Training}
\label{sec:analysis_training}

%\paragraph{Training Efficiency}
%In reality, training speed is important in machine translation with many directions.
%\begin{table}[h]
%\centering
%\small
%\begin{tabular}{l|cccc}
%\hline
%& \multicolumn{2}{c}{\textbf{en$\rightarrow$de}} & \multicolumn{2}{c}{\textbf{zh$\rightarrow$en}} \\
%Training Variant & \\
%AT & \\
%CMLM & \\
%Multimask & \\
%Levenshtein Transformer & \\
% \hline
%\end{tabular}
%\caption{Training time of 300K updates on 16 Tesla V100 GPUs.}
%\label{tab:training_time}
%\end{table}

%\begin{figure}
%\centering
%    \includegraphics[width=0.46\textwidth]{val_bleu.pdf}
%\label{val_bleu}
%\caption{En$\rightarrow$fr validation results of the multimask and CMLM over training updates.}
%\end{figure}

\begin{figure}[t]
\centering
    \includegraphics[width=0.47\textwidth]{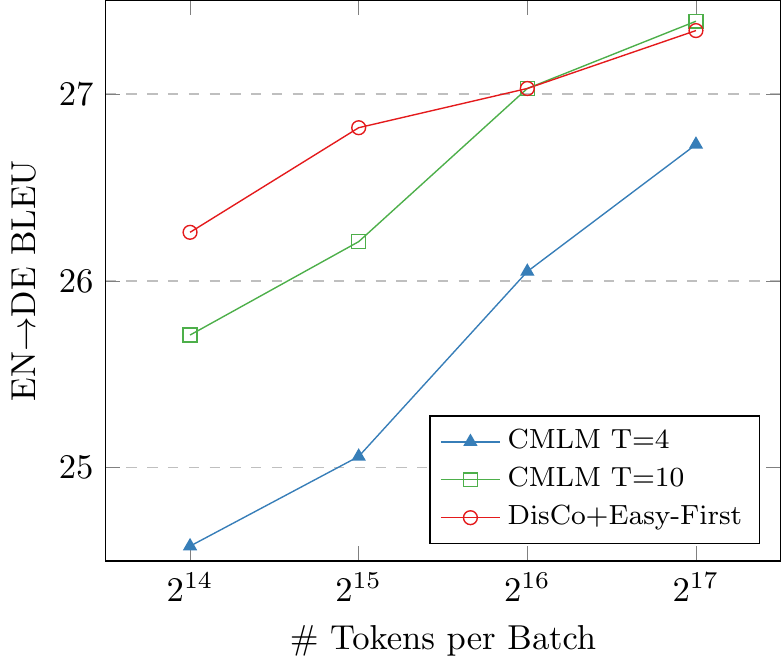}
\caption{EN$\rightarrow$DE test results with varying batch size.}
\label{fig:training_efficiency}
\end{figure}
\paragraph{Training Efficiency}
In Sec.\ \ref{subsec:generalization}, we discussed the fundamental inefficiency of CMLM training---a CMLM model is trained to only predict a subset of the target words. DisCo addresses this problem by its architecture that allows for predicting every word given a randomly chosen subset of the target words. Seen in Fig.\ \ref{fig:training_efficiency} are results on the en$\rightarrow$de test data with varying batch sizes. We can see that DisCo is more robust to smaller batch sizes, supporting our claim that it provides more efficient training.

\paragraph{Distillation}
We assess the effects of knowledge distillation across different models and inference configurations (Table \ref{tab:distill}).
Consistent with previous models \cite{Gu2017NonAutoregressiveNM, UnderstandingKD}, we find that distillation facilitates all of the non-autoregressive models.
Moreover, the DisCo transformer benefits more from distillation compared to the CMLM under the same mask-predict inference.
This is in line with \citet{UnderstandingKD} who showed that there is correlation between the model capacity and distillation data complexity.
The DisCo transformer uses contextless keys and values, resulting in reduced capacity.
Autoregressive translation also improves with distillation from a large transformer, but the difference is relatively small.
Finally, we can observe that the gain from distillation decreases as we incorporate more global information in inference (more iterations in NAT cases and larger beam size in AT cases).

\begin{table}[h]
\centering
\small
\caption{Effects of distillation across different models and inference. All results are BLEU scores from the dev data. $T$ and $b$ denote the max number of iterations and beam size respectively.}
\vskip 0.1in
\addtolength{\tabcolsep}{-1.0pt}  
\begin{tabular}{l|c|ccc|ccc}
\toprule
&  & \multicolumn{3}{c|}{\textbf{en$\rightarrow$de}} & \multicolumn{3}{c}{\textbf{ro$\rightarrow$en}} \\
Model & $T$ & raw\ & dist. &$\Delta$ & raw\ & dist. & $\Delta $\\
 \hline
CMLM + MaskP& 4 &22.7  & 25.5& 2.8 & 33.2 &  34.8 & 1.6\\
CMLM + MaskP & 10 & 24.5& 25.9 & 1.4 & 34.5 &  34.9 & 0.4 \\
\hdashline
DisCo + MaskP & 4& 21.4 &24.6 & 3.2 &   32.3 & 34.1 & 1.8\\
DisCo + MaskP & 10 & 23.6 & 25.3 & 1.7 & 33.4 & 34.3 & 0.9 \\ 
%Mmask + MaskP & 20 & 24.1 & 25.7 & 1. morning  &  & 3 & \\ 
\hdashline
DisCo + EasyF& 10 & 23.9 &  25.6 & 1.7 &   34.0&35.0 & 1.0 \\ \hline
AT Base ($b=1$)& N & 25.5 & 26.4  & 0.9 & -- & --& -- \\ 
%AT Base ($b=3$)& N & 25.9& 26.5& 0.6 &  -- & -- & -- \\
AT Base ($b=5$)& N & 26.1 &26.8 & 0.7 &  -- & -- & -- \\
%CMLM + MaskP & 4 &22.26 & 24.83& 33.16 &  34.70 \\
%CMLM + MaskP & 10 & 24.22& 25.23 & 34.48 &  35.16\\
%CMLM + MaskP & 10 & 24.2& 25.2 & 34.5 &  35.2\\
%\hdashline
%Mmask + MaskP & 4& 21.37 &24.58 &  30.78 &  34.06\\
%Mmask + MaskP & 10 & 23.63 & 25.34 & 32.33  & 34.27 \\
%\hdashline
%Mmask + EasyF& 10 & 23.92 &  25.60 & 32.17 &34.46  \\ \hline
%AT Base ($b=1$)& N & 25.24 &  26.23 &  -- & --  \\
%AT Base ($b=5$)& N & 25.89 & 26.43  &  -- & --  \\
\bottomrule
\end{tabular}
\label{tab:distill}
\end{table}

\paragraph{AT with Contextless KVs}
We saw that a decoder with contextless keys and values can still retain performance in non-autoregressive models. Here we use a decoder with contextless keys and values in autoregressive models. The results (Table \ref{tab:contextless_kv}) show that it is able to retain performance even in autoregressive models regardless of distillation, suggesting further potential of our approach.
\begin{table}[h]
\centering
%\small
\caption{Test results (BLEU) from AT with contextless keys and values.}
\vskip 0.1in
\begin{tabular}{l|ccccc}
\hline
AT & \multicolumn{2}{c}{\textbf{en$\rightarrow$de}} & \multicolumn{2}{c}{\textbf{de$\rightarrow$en}}& \multicolumn{1}{c}{\textbf{ro$\rightarrow$en}} \\
Decoder & raw & dist.\ & raw & dist.\ & raw\\
 \hline
Contextless &  27.09& 27.86 & 30.91 & 31.46 & 34.25\\
Original &  26.85 & 27.69 & 31.33 & 31.09 & 34.46\\
\hline
\end{tabular}
\label{tab:contextless_kv}
\vskip -0.05in
\end{table}

\paragraph{Easy-First Training}
So far we have trained our models to predict every word given a random subset of the other words.
But this training scheme yields a gap between training and inference, which might harm the model.
We attempt to make training closer to inference by training the DisCo transformer in the easy-first order.
Similarly to the inference, we first predict the easy-first order by estimating $P(Y_n|X)$ for all $n$. Then, use that order to determine $Y_{\obs}^{n}$.\footnote{This training process can be seen as the hard EM algorithm where the easy-first order is a latent variable.}
The overall loss will be the sum of the negative loglikelihood of these two steps.
Seen in Table \ref{tab:training_strategies} are the results on the dev sets of en$\rightarrow$de and ro$\rightarrow$en.
In both directions, this easy-first training does not ameliorate performance, suggesting that randomness helps the model.
Notice also that the average number of iterations in inference decreases (4.03 vs.\ 4.29, 2.94 vs.\ 3.17).
The model gets trapped in a sub-optimal solution with reduced iterations due to lack of exploration.

\begin{table}[h]
\centering
%\small
\caption{Dev results from bringing training closer to inference.}
\vskip 0.1in
\begin{tabular}{l|cccc}
\hline
& \multicolumn{2}{c}{\textbf{en$\rightarrow$de}} & \multicolumn{2}{c}{\textbf{ro$\rightarrow$en}} \\
Training Variant & Step\ & BLEU  & Step\ & BLEU\\
 \hline
Random Sampling & 4.29 &  \textbf{25.60} & 3.17  & \textbf{34.97}\\
Easy-First Training & 4.03 &  24.76 & 2.94 & 34.96 \\
\hline
\end{tabular}
\label{tab:training_strategies}
\vskip -0.05in
\end{table}

\begin{table}
\centering
%\small
\vskip -0.15in
\caption{Dev results with different decoding strategies.}
\vskip 0.1in
\begin{tabular}{l|cccc}
\toprule
& \multicolumn{2}{c}{\textbf{en$\rightarrow$de}} & \multicolumn{2}{c}{\textbf{ro$\rightarrow$en}} \\
\textbf{Inference Strategy}& Step\ & BLEU& Step\ & BLEU \\
\hline
Left-to-Right Order & 6.80 &21.25 & 4.86 & 33.87\\
Right-to-Left Order & 6.79 &20.75 & 4.67 & 34.38\\
All-But-Itself  & 6.90 &  20.72 & 4.80  &  33.35 \\ \hline
Parallel Easy-First & 4.29 & \textbf{25.60}&  3.17 & \textbf{34.97} \\
Mask-Predict &10 & 25.34 & 10  & 34.54 \\
\bottomrule
\end{tabular}
\label{tab:decoding_strategies}
\end{table}

\subsection{Inference}
\label{sec:analysis_inference}

%\paragraph{Iterations}

\begin{figure*}[t]
    \centering
    \includegraphics[width=0.998\textwidth]{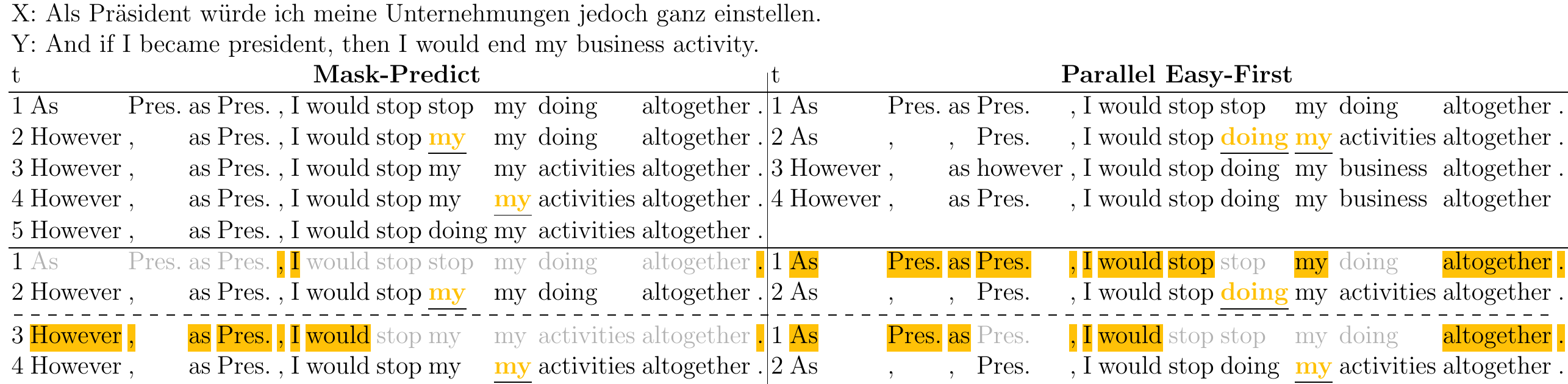}
    \caption{An example of inference iterations in de$\rightarrow$en from the dev set when max iteration $T$ is 5. (\textit{Pres.}\ stands for \textit{President}). We show how each of the \tyellow{underscored words} were generated in the bottom section. Prediction is conditioned on \hly{highlighted tokens}.}
\label{fig:example_deen}
\end{figure*}
\begin{figure}[t]
\centering
    \includegraphics[width=0.47\textwidth]{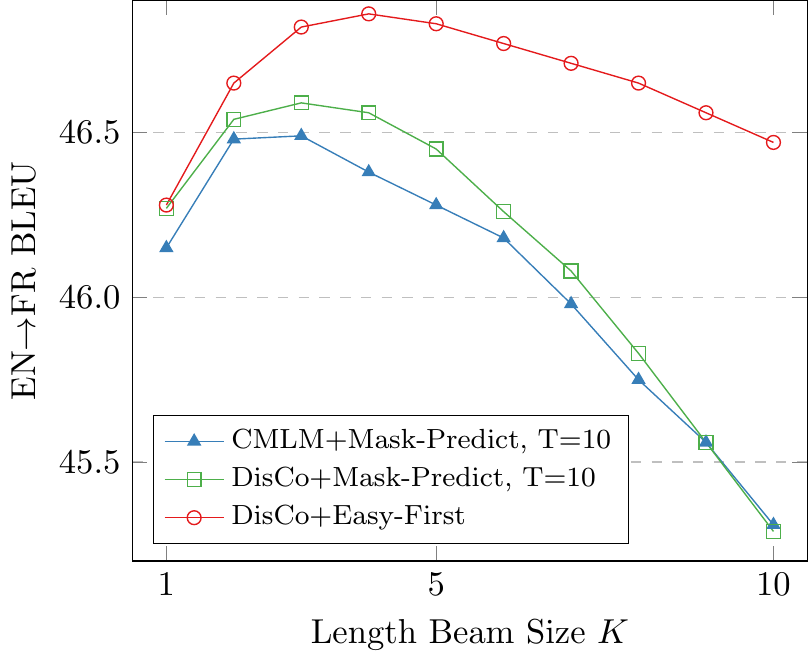}
\caption{EN$\rightarrow$FR dev results with varying length beam size.}
\label{fig:length_beam}
\end{figure}
\paragraph{Alternative Inference Algorithms}
Here we compare various decoding strategies on the DisCo transformer (Table \ref{tab:decoding_strategies}).
Recall in the parallel easy-first inference (Sec.\ \ref{sec:easy-first}), we find the easy-first order by sorting the probabilities in the first iteration and compute each position's probability conditioned on the \textit{easier} positions from the previous iteration.
We evaluate two alternative orderings: left-to-right and right-to-left.
We see that both of them yield much degraded performance.
We also attempt to use even broader context than parallel easy-first by computing the probability at each position based on all other positions (\textit{all-but-itself}, $Y_{obs}^{n,t} = Y_{\neq n}^{t-1}$).
We again see degraded performance, suggesting that cyclic dependency (e.g.,\ $Y_{m}^{t-1} \in Y_{\obs}^{n,t}$ and $Y_{n}^{t-1} \in Y_{\obs}^{m, t}$) breaks consistency.
For example, a model can have two output candidates: ``Hong Kong" and ``New York" \cite{Zhang2020POINTERCT}. In this case, we might end up producing ``Hong York" due to this cyclic dependency.
These results suggest that the easy-first ordering we introduced is a simple yet effective approach.

%We first compare the easy-first order with the left-to-right and right-to-left order.
%We see that left-to-right and right-to-left significantly yield significantly degraded performance.
%We also perform the all-but-itself inference.
%Notice that if we take $T=N$ where $N$ indicates the length of the target sentence, then the left-to-right method reduces to autoregressive decoding of the multimask transformer, constrainted by predicted length.

\paragraph{Example Translation}
Seen in Fig.\ \ref{fig:example_deen} is a translation example in de$\rightarrow$en when decoding the same DisCo transformer with the mask-predict or parallel easy-first inference.
In both algorithms, iterative refinement resolves structural inconsistency, such as repetition.
Parallel easy-first succeeds in incorporating more context in early stages whereas mask-predict continues to produce inconsistent predictions (``my my activities") until more context is available later, resulting in one additional iteration to land on a consistent output.

\paragraph{Length Beam}
Fig.\ \ref{fig:length_beam} shows performance of the CMLM and DisCo transformer with varying size of length beam.
All cases benefit from multiple candidates with different lengths to a certain point, but DisCo + Easy-First improves most.
This can be because parallel easy-first relies on the easy-first order as well as the length, and length beam provides opportunity to try multiple orderings.

\begin{figure}[t]
\centering
    \includegraphics[width=0.47\textwidth]{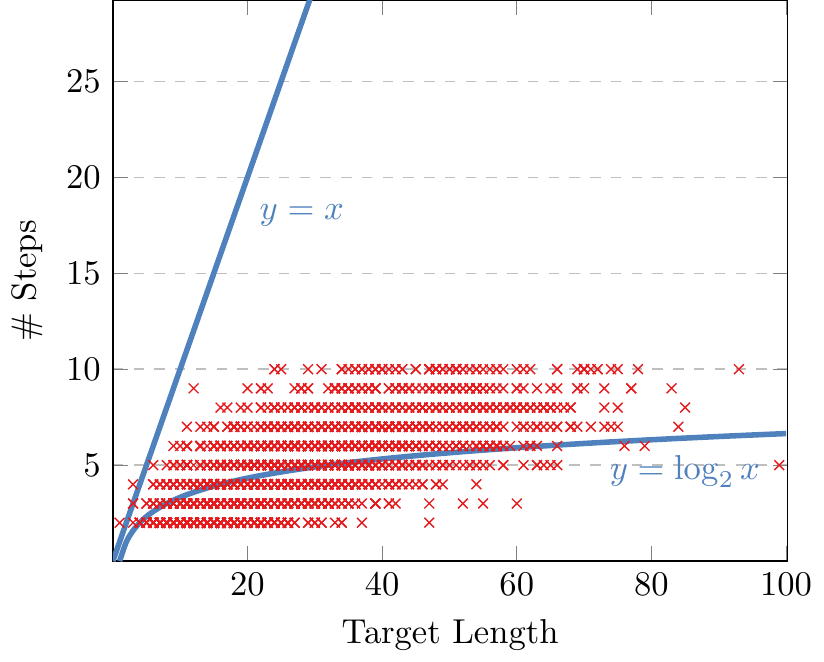}
\caption{\# Refinement steps vs.\ target length on the WMT14 en$\rightarrow$de test data.}
\label{fig:length_iter}
\end{figure}

\paragraph{Iterations vs.\ Length}
We saw that parallel easy-first inference substantially reduced the number of required iterations.
We hypothesize that the algorithm effectively adapts the number of iterations based on the difficulty, which is reminiscent of a dynamic halting mechanism \cite{Graves2016AdaptiveCT, Dehghani2019UniversalT}.
To see this, we compare the number of required iterations and the generated sentence length as a proxy for difficulty (Fig.\ \ref{fig:length_iter}).
Similarly to the experiments above, we set the max iteration and length beam to be 10 and 5 respectively.
While the number of required iterations vary to a certain degree, we see that long sentences tend to require more iterations to converge.

\section{Related and Future Work}
\label{sec:related_work}
In addition to the work discussed above, prior and concurrent work on non-autoregressive translation developed ways to mitigate the trade-off between decoding parallelism and performance. 
As in this work, several prior and concurrent work proposed methods to iteratively refine (or insert) output predictions \cite{Mansimov2019AGF, Stern2019InsertionTF, Gu2019InsertionbasedDW, Chan2019KERMITGI, Chan2019AnES, Ghazvininejad2020SemiAutoregressiveTI, Li2020LAVANA, Saharia2020}.
Other approaches include adding a lite autoregressive module to parallel decoding \cite{Kaiser2018FastDI,Sun2019Fast,ReorderNAT}, partially decoding autoregressively \cite{Stern2018BlockwisePD,Stern2019InsertionTF}, rescoring output candidates autoregressively (e.g.,\ \citealp{Gu2017NonAutoregressiveNM}), mimicking hidden states of an autoregressive teacher \cite{Li2019HintBasedTF}, training with different objectives than vanilla negative log likelihood \cite{Libovick2018EndtoEndNN,Wang2019NonAutoregressiveMT,minBOW, Ghazvininejad2020AlignedCE, Li2020LAVANA}, reordering input sentences \cite{ReorderNAT}, generating with  an energy-based inference network \cite{Tu2020ENGINEEI}, training on additional data from an autoregressive model \cite{Zhou2020ImprovingNN}, and modeling with latent variables \cite{Ma2019FlowSeqNC, Shu2019LatentVariableNN}. 

While this work took iterative decoding methods, our DisCo transformer can be combined with other approaches for efficient training.
For example, \citet{Li2019HintBasedTF} trained two separate non-autoregressive and autoregressive models, but it is possible to train a single DisCo transformer with both autoregressive and random masking and use hidden states from autoregressive masking as a teacher.
%our transformer can be trained to be both autoregressive and and non-autoregressive and hidden states can b. 
%For example, the Levenshtein transformer training involves deleting words randomly from the ground truth and feeding each to the transformer, but our decoder can simulate multiple instances of deletion in one pass.
We leave integration of the DisCo transformer with more approaches to non-autoregressive translation for future.

We also note that our DisCo transformer can be used for general-purpose representation learning.
In particular, \citet{Liu2019RoBERTaAR} found that masking different tokens in every epoch outperforms static masking in BERT \cite{devlins2019bert}.
Our DisCo transformer would allow for making a prediction at every position given arbitrary context, providing even more flexibility for large-scale pretraining.
%Moreover, we can reduce training time significantly by contextless computation, which benefits large-scale self-supervised pretraining.

%\subsection{Architecture}
%Tying decoding layers.
%it becomes more natural to tie layers. 

\section{Conclusion}
We presented the DisCo transformer that predicts every word in a sentence conditioned on an arbitrary subset of the other words. 
We developed an inference algorithm that takes advantage of this efficiency and further speeds up generation without loss in translation quality.
Our results provide further support for the claim that non-autoregressive translation is a fast viable alternative to autoregressive translation. 
Nonetheless, a discrepancy still remains between autoregressive and non-autoregressive performance when knowledge distillation from a large transformer is applied to both.
We will explore ways to narrow this gap in the future.

% Acknowledgements should only appear in the accepted version.
\section*{Acknowledgements}
We thank Tim Dettmers, Hao Peng, Mohammad Rasooli, William Chan, and Qinghong Han as well as the anonymous reviewers for their helpful feedback on this work.

% In the unusual situation where you want a paper to appear in the
% references without citing it in the main text, use \nocite
\nocite{langley00}

\bibliography{example_paper}
\bibliographystyle{icml2020}

\end{document}